%% file: main.tex
\documentclass{article}  
\usepackage[final]{nips_2017}
\usepackage[utf8]{inputenc} 
\usepackage[T1]{fontenc}    
\usepackage{hyperref}       
\usepackage{url}            
\usepackage{booktabs}       
\usepackage{amsfonts}    
\usepackage{color}
\usepackage{epsfig}
\usepackage[linesnumbered,ruled,vlined]{algorithm2e}
\usepackage{amsthm}
\usepackage{amsmath}
\usepackage{float}
\usepackage{graphicx}
\usepackage{wrapfig}
\usepackage{makecell}
\usepackage{caption}
\usepackage{subcaption}

\newcommand{\nop}[1]{}
\DeclareMathOperator*{\argmax}{arg\,max}

\newtheorem{theorem}{Theorem}

\newtheorem{lemma}[theorem]{Lemma}

\title{Learning K-way D-dimensional Discrete Code For Compact Embedding Representations}

\author{
	Ting Chen\\
	UCLA \\
	\texttt{tingchen@cs.ucla.edu} \\
	\And
	Martin Renqiang Min\\
	NEC Labs America \\
	\texttt{renqiang@nec-labs.com} \\
	\And
	Yizhou Sun\\
	UCLA \\
	\texttt{yzsun@cs.ucla.edu} \\
}

\begin{document}

\maketitle

\input{content/abstract}
\input{content/introduction}

\input{content/model}

\input{content/theory}
\input{content/exp}
\input{content/related}
\input{content/conclusion}

\small
\bibliographystyle{unsrt}
\bibliography{content/ref}

\end{document}

%% file: content/abstract.tex
\begin{abstract}	
	Embedding methods such as word embedding have become pillars for many applications containing discrete structures. Conventional embedding methods directly associate each symbol with a continuous embedding vector, which is equivalent to applying linear transformation based on ``one-hot'' encoding of the discrete symbols. Despite its simplicity, such approach yields number of parameters that grows linearly with the vocabulary size and can lead to overfitting. In this work we propose a much more compact K-way D-dimensional discrete encoding scheme to replace the ``one-hot" encoding. In ``KD encoding'', each symbol is represented by a $D$-dimensional code, and each of its dimension has a cardinality of $K$. The final symbol embedding vector can be generated by composing the code embedding vectors. To learn the semantically meaningful code, we derive a relaxed discrete optimization technique based on stochastic gradient descent. By adopting the new coding system, the efficiency of parameterization can be significantly improved (from linear to logarithmic), and this can also mitigate the over-fitting problem. In our experiments with language modeling, the number of embedding parameters can be reduced by 97\% while achieving similar or better performance.
\end{abstract}

%% file: content/introduction.tex
\section{Introduction}

Embedding methods, such as word embedding \cite{mikolov2013distributed,pennington2014glove}, have become pillars in many applications when learning from discrete structures. The examples include language modeling \cite{kim2016character}, machine translation \cite{sennrich2015neural}, text classification \cite{zhang2015character}, knowledge graph and social network modeling \cite{bordes2013translating}, and many others \cite{chen2016entity}. The objective of the embedding module in neural networks is to represent a discrete symbol, such as a word or an entity, with some continuous embedding vector $v\in R^{d}$. This seems to be a trivial problem, at the first glance, in which we can directly associate each symbol with a learnable embedding vector, as it is done in existing work. To retrieve the embedding vector of a specific symbol, an embedding table lookup operation can be performed. This is equivalent to the following: first we encode each symbol with an ``one-hot'' encoding vector $b \in [0,1]^N$ where $\sum_j b_j=1$ ($N$ is the total number of symbols); then to generate the embedding vector, we simply multiply the ``one-hot'' vector $b$ with the embedding matrix $W\in R^{N\times d}$, i.e. $v=b^TW$.

Despite the simplicity of this ``one-hot'' encoding based embedding approach, it has several issues. The major issue is that the number of parameters grows linearly with the number of symbols. This becomes very challenging when we have millions or billions of entities in the database, or when there are lots of symbols with only a few observations (e.g. Zipf's law). There also exists redundancy in the $O(N)$ parameterization, assuming many symbols may actually be similar to each other. This over-parameterization can further lead to overfitting; and it also requires a lot of memory, which prevents the model from being deployed to mobile devices. Another issue is purely from the code space utilization perspective, where we find ``one-hot'' encoding is extremely inefficient. Its code space utilization rate is almost zero as $N / 2^N \rightarrow 0$, while $N$ bits/dimensions of code can effectively represent $2^N$ symbols.

To address these issues, we propose a novel and much more compact coding scheme that replaces the ``one-hot'' encoding. In the proposed approach, we use a $K$-way $D$-dimensional code to represent each symbol, where each code has $D$ dimensions, and each dimension has a cardinality of $K$. For example, a concept of cat may be encoded as (5-1-3-7), and a concept of dog may be encoded as (5-1-3-9). The code allocation for each symbol is based on data such that they will be able to capture semantics of symbols, and similar codes may reflect similar meanings. We dub the proposed encoding scheme as ``\textit{KD encoding}''.

The KD code system is much more compact than its ``one-hot'' counterpart. To represent a set of symbols of size $N$, the ``KD encoding'' only requires that $K^D\ge N$. By increasing $K$ or $D$ by a small amount, we can easily achieve $K^D \gg N$, in which case it will still be much more compact. Consider $K=2$, the utilization rate of ``KD encoding'' is $N/2^D$, which is $2^{N-D}$ times more compact than ``one-hot'' counterpart \footnote{Assuming we have vocabulary size $N=10,000$, and setting number of dimensions $D=100$, that is $2^{9900}$ times more efficient}.

The compactness of the code can be translated into compactness of the parametrization. Dropping the giant embedding matrix $W\in R^{N\times d}$ that stores symbol embeddings, the symbol embedding vector is generated by composing much fewer code embedding vectors. This can be achieved as follows: first we embed each KD code into a sequence of vector in $R^{D\times d'}$, and then apply some transformation $f(\cdot)$, which can be based on neural networks, to generate the final symbol embedding. In order to learn meaningful discrete codes that can exploit the similarities among symbols, we derive a relaxed discrete optimization algorithm based on stochastic gradient descent (SGD). By adopting the new approach, we can reduce the the number of parameters form $O(Nd)$ to $O(\frac{K}{\log K} d' \log N + C)$, where $d'$ is the code embedding size, and $C$ is the number of neural network parameters. To validate our idea, we conduct experiments on both synthetic data as well as a real language modeling task. We achieve 97\% of embedding parameter reduction in the language modeling task and obtain similar or better performance.

%% file: content/model.tex
\section{The K-way D-dimensional Discrete Encoding}

In this section we introduce the ``KD encoding'' in details. Specifically, we present methods to generate symbol embedding from its (given/learned) ``KD code'', and also the techniques for learning ``KD code'' from the data.

\subsection{The ``KD encoding'' Framework}
In the proposed framework, each symbol is associated with a $K$-way and $D$-dimensional discrete code. We denote each symbol by $s\in \mathcal{S}$, where $\mathcal{S}$ is a set of symbols with cardinality $N$. And each discrete code is denoted by $c_i = (c_i^1, c_i^2, \cdots, c_i^D) \in \mathcal{B}^D$, where $\mathcal{B}$ is the set of code bits with cardinality $K$. To connect symbols with discrete codes, a mapping function $\phi(\cdot): \mathcal{S} \rightarrow \mathcal{B}^D$ is used. The learning of this mapping function will be introduced later, and once fixed it can be stored as a hash table for fast lookup.

\begin{figure}[]
	\begin{center}
		\includegraphics[trim=0 180 50 190,clip,width=\textwidth]{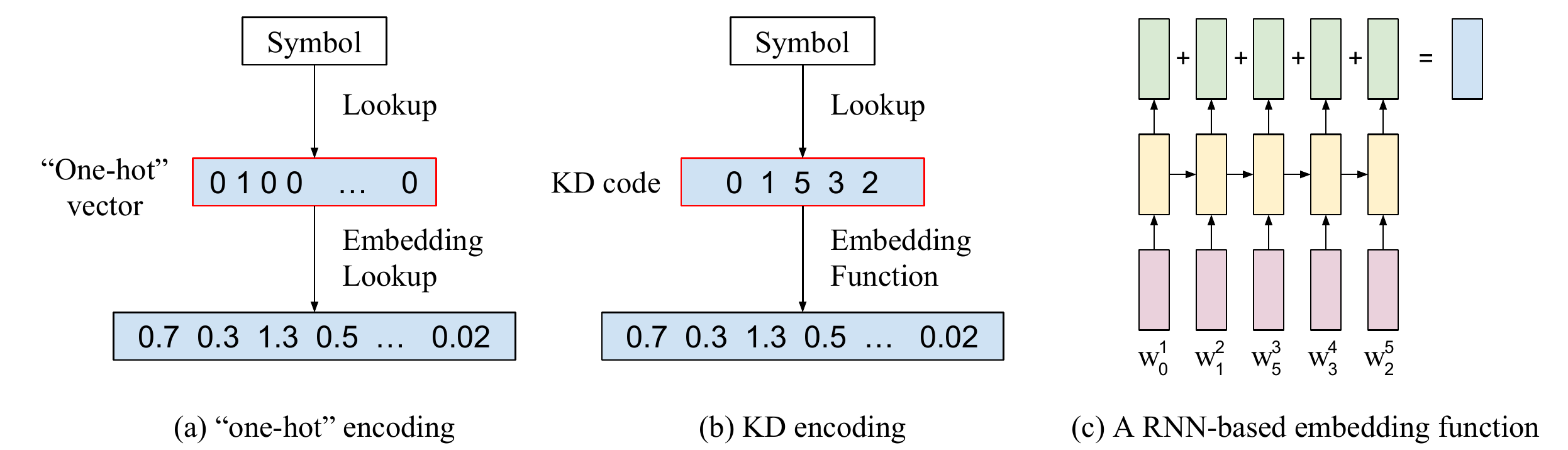}
	\end{center}
	\caption{\label{fig:framework} (a) The conventional symbol embedding based on ``one-hot'' encoding. (b) The proposed KD encoding scheme. (c) An example of embedding transformation function by RNN used in the KD encoding when generating the symbol embedding from the code.}

\end{figure}

Given the $i$-th symbol $s_i$, we can retrieve its code via a code lookup, $c_i = \phi(s_i)$. The final embedding $v$ is generated by first embedding the code $c_i$ to a sequence of code embedding vectors $(\mathcal{W}^1_{c_i^1}, \mathcal{W}^2_{c_i^2}, \cdots, \mathcal{W}^D_{c_i^D})$, and then apply a differentiable transformation function $v = f(\mathcal{W}^1_{c_i^1}, \mathcal{W}^2_{c_i^2}, \cdots, \mathcal{W}^D_{c_i^D}; \theta)$, which is learned as well. We introduce the transformation function $f(\cdot)$ in the next sub-section. Here $\mathcal{W}^j\in R^{K\times d'}$ is the embedding matrix for the $j$-th code bit. The overall framework is illustrated in Figure \ref{fig:framework}.

In order to uniquely identify a symbol, we only need that $K^D = N$, as we can assign an unique code to each symbol. When this holds, the code space is fully utilized, and none of the symbol can change its code without affecting the other symbols. We call this type of code system the \textit{compact code}. The optimization problem for compact code can be very difficult, and usually requires approximated combinatorial algorithms such as graph matching \cite{li2016lightrnn}. Opposite to the compact code is the \textit{redundant code} system, where we have $K^D \gg N$, and there will be a lot of ``empty'' code space that has no symbol correspondence, so that changing the code of one symbol may not affect other symbols, since the random collision probability can be very small \footnote{For example, we can set $K=100, D=10$ for a billion symbols, in a random code assignment, the probability of the NO collision at all is 99.5\%.}, which makes it easier to optimize. The redundant code can be achieved by slightly increasing the size of $K$ or $D$ thanks to the exponential nature of their relations. Hence, in both compact code or redundant code, we have $D = O(\frac{\log N}{\log K})$.

\subsection{Discrete Code Embedding}
Since a discrete code has multiple bits/dimensions, we cannot directly use embedding lookup to find the symbol embedding as used in ``one-hot'' encoding. Hence, we first map each code into code embedding vectors via a code lookup $c_i = \phi(s_i)$, and then use a function $f(\cdot)$ that transforms the code embedding vectors into the final symbol embedding vector.

As mentioned above, we associate an embedding matrix $\mathcal{W}^j \in R^{K\times d'}$ for each $j$-th dimension in the discrete code. this enables us to turn a discrete code $c_i$ into a sequence of code embedding vectors $(\mathcal{W}^1_{c_i^1}, \mathcal{W}^2_{c_i^2}, \cdots, \mathcal{W}^D_{c_i^D})$. 

Now to generate the final embedding vector $v$, a transformation function $f(\cdot)$ is applied. In this work we consider two types of embedding transformation functions. The first one is based on a linear transformation,
$$
v_i = \bigg(\sum_j \mathcal{W}^j_{c_i^j}\bigg)^TH
$$
Where $H\in R^{d'\times d}$ is the linear matrix. While this is simple, due to its linear nature, the capacity of the generated symbol embedding can be limited. This motivates us to adopt a non-linear transformation function based on a recurrent neural network, LSTM \cite{hochreiter1997long}, in particular. Assuming the code embedding dimension is the same as the LSTM hidden dimension, the formulation is given as follows.
\begin{gather*}
f_j = \sigma(\mathcal{W}^j_{c^j} + U_{f} h_{j-1} + b_f) \\
i_j = \sigma(\mathcal{W}^j_{c^j} + U_{i} h_{j-1} + b_i) \\
o_j = \sigma(\mathcal{W}^j_{c^j} + U_{t} h_{j-1} + b_t) \\
m_j = f_j \circ m_{j-1} + i_j \circ \tanh(\mathcal{W}^j_{c^j}+ U_{m} h_{j-1} + b_m) \\
h_j = o_j \circ \tanh(m_j),
\end{gather*}
where $\sigma(\cdot)$ and $\tanh(\cdot)$ are, respectively, standard sigmoid and tanh activation functions. Please also noted the symbol index $i$ is ignored for simplicity. The final symbol embedding can be computed by summing over LSTM outputs at all code bits (with a linear transformation to match dimension if $d\ne d'$), i.e. $v = (\sum_j h_j)^TH$.
\begin{lemma}
The number of embedding parameters used in KD encoding is $O(\frac{K}{\log K} d' \log N + C)$, where $C$ is the number of parameters of neural nets.
\end{lemma}
The proof is straight-forward. There are two types of embedding parameters in the KD encoding: (1) code embedding vectors, and (2) neural network parameters. And there are $O(\frac{K}{\log K} \log N)$ code embedding vectors with $d'$ dimensions. As for the number of parameters in neural networks (LSTM) $C$ that is in $O(d'^2)$, it may be treated as a constant to the number of symbols since $d'$ is independent of $N$, provided that there are certain structures presented in the symbol embeddings. For example, if we assume the symbol embeddings are within $\epsilon$-ball of a finite number of centroids in $d$-dimensional space, it should only require a constant $C$ to achieve $\epsilon$-distance error bound, regardless of the vocabulary size, since the neural networks just have to memorize the finite centroids.

\subsection{Discrete Code Learning}

The code assignment can be very important for both parameterization efficiency and generalization. So we want to learn the code allocation function $\phi(\cdot): s \rightarrow c$ end-to-end from data, in contrast to hand-coded ``one-hot'' encoding. In this work, we assume that we are already given the pre-trained embedding vectors $\mathbf{v} = (v_1, v_2, \cdots, v_N)$ and each $v_i \in R^d$. Thus we will learn the discrete codes based on given $\mathbf{v}$. Once the codes are learned, we can re-learn the code embedding parameters including transformation function $f(\cdot)$ according to the specific task. In the future, we will extend it to the case where such embeddings are not available.

To find the optimal codes, we minimize the squared loss between the real embedding vector $v_i$ and the embedding vector generated from the KD code. This yields to the following.
\begin{equation}
\min_{\theta, \{\mathcal{W}\}, \{c_i\}} \sum_i \bigg(v_i - f\bigg(\mathcal{W}^1_{c_i^1}, \mathcal{W}^2_{c_i^2}, \cdots,\mathcal{W}^D_{c_i^D}; \theta\bigg)\bigg)^2
\end{equation}
Where $f$ is a differentiable transformation function as introduced above.

\begin{wrapfigure}{r}{0.24\textwidth}
	\begin{center}
	\includegraphics[width=0.24\textwidth]{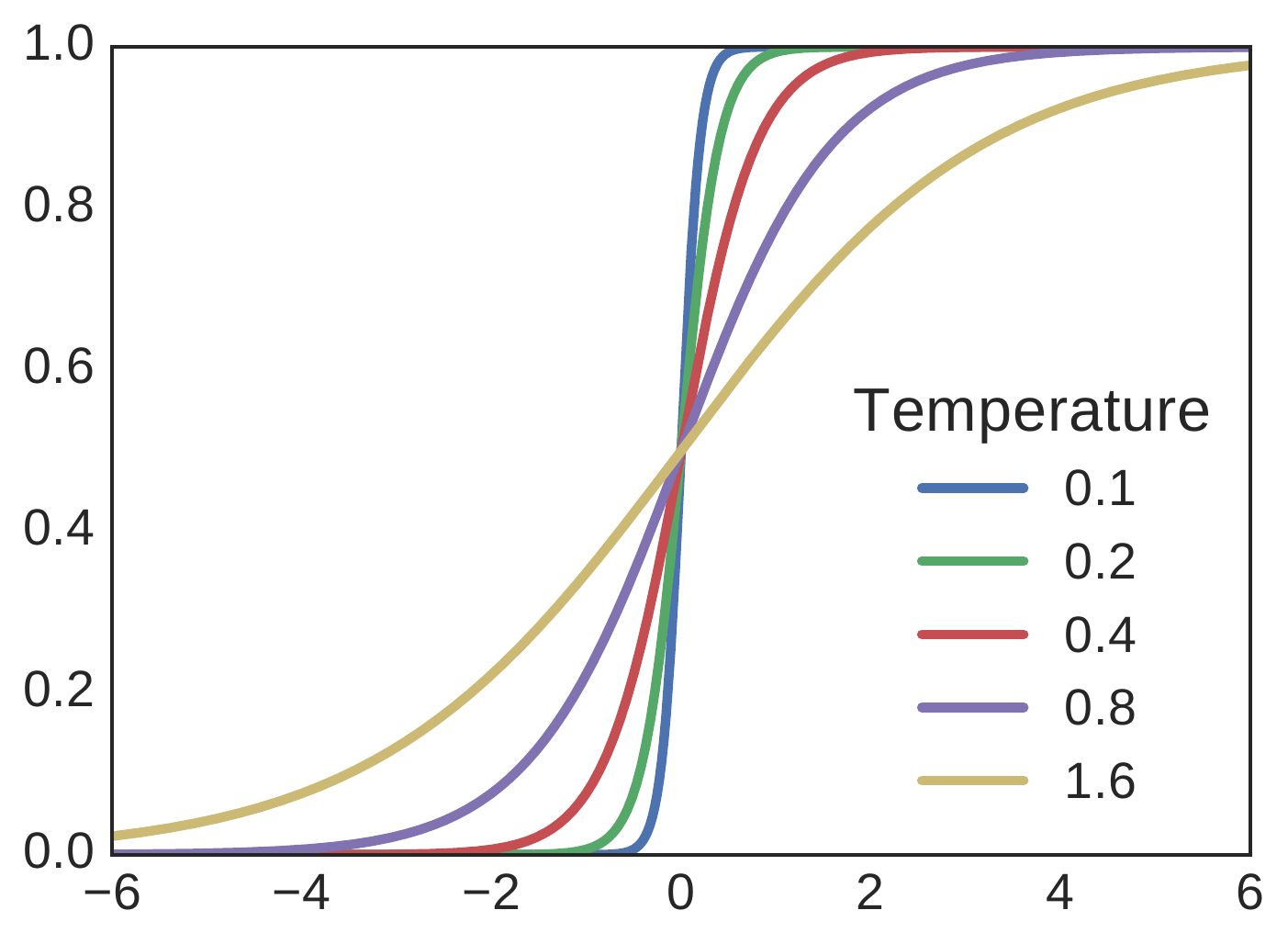}
	\end{center}
	\caption{\label{fig:sigmoid_wtemp} The effects of temperature ($K=2$).}
\end{wrapfigure}

Since each $c$ is a discrete code, it cannot be directly optimized via stochastic gradient descent as other parameters do. Thus we need to use a relaxation in order to learn it effectively via SGD. We observe that each code $c_i$ can be seen as a concatenation of $D$ ``one-hot'' vector, i.e. $c_i = (o^{1}_{i}, o^{2}_i, \cdots, o^{D}_i)$, where $\forall j, o^j_i\in [0, 1]^K$ and $\sum_k o^{jk}_i = 1$, where $o^{jk}_i$ is the $k$-th component of $o^{j}_i$. We can adjust $o^j_i$ in order to update the code, but it is still non-differentiable. To address the issue, we relax the $o^j_i$ from an ``one-hot'' vector to some continuous vector by applying \textit{tempering Softmax}:
$$
o^{jk}_i \approx \frac{\exp(\hat{o}^{jk}_i/T)}{\sum_{k'} \exp(\hat{o}^{jk'}_i/T)}
$$
Where $T$ is a temperature term, as $T\rightarrow 0$, this approximation becomes exact (except for the case of ties). Similar techniques have been applied in Gumbel-Softmax \cite{jang2016categorical,maddison2016concrete}. We show effects of the temperature when $K=2$ with $y = 1/(1+\exp(-x/T))$ in Figure \ref{fig:sigmoid_wtemp}.

To learn the relaxed code logits $\hat{o}^j_i$, we can gradually decrease the temperature $T$ during the training. When $T$ is not small enough, $o^j_i$ is still a smooth vector, so we use linear combination, i.e. $(o^j_i)^T \mathcal{W}^j$, instead of indexing, i.e. $\mathcal{W}^j_{c^j_i}$, to generate the embedding vector for $j$-th code dimension.

Noted that the tempering Softmax approximation is only differentiable when $T$ is not too small, but the gradient will disappear when $T\rightarrow 0$. So at the beginning when $T$ is not small enough, we are actually learning some continuous codes instead of discrete codes, which may not be desirable. When $T$ becomes small enough such that we start to learn real discrete codes, the small $T$ in turn prevents the code from further update as it makes gradient disappear.

To address this issue, we take inspiration from Straight-Through Estimator \cite{bengio2013estimating}. In the forward pass, instead of using the tempering Softmax output, which is likely a smooth continuous vector, we take its maximum and turn it into a ``one-hot'' vector as follows, which resembles the exactly discrete code.
$$
o^j_i = \text{one\_hot}\bigg(\argmax_k \hat{o}^{jk}_i\bigg) \approx \text{Softmax}\bigg(\frac{\hat{o}^j_i}{\epsilon}\bigg)\text{,~~~~~}\epsilon\rightarrow 0
$$
The use of straight-through estimator is equivalent to use different temperatures during the forward and backward pass. In forward pass, $T\rightarrow 0$ is used, for which we simply take the argmax. In the backward pass (to compute the gradient), we pretend that a larger $T$ was used. Although this is a biased gradient estimator, but the sign of the gradient is still correct. Compared to using the same temperatures in both passes, this always output ``one-hot'' discrete code $o^j_i$, and there is no vanishing gradient problem as long as the backward temperature is not approaching zero.

The training procedure is summarized in Algorithm \ref{algo:STE_softmax}, in which the \texttt{stop\_gradient} operator will prevent the gradient from back-propagating through it.

\begin{algorithm}
	\small
	\DontPrintSemicolon 
	\KwIn{Symbol embedding $\{v_i\}$, code logits $\{\hat{o}_i\}$, code embedding matrices $\{\mathcal{W}^j\}$, transformation parameters $\theta$.}
	\KwOut{Discrete codes \{$o_i$\}.}
	\For{$i \gets 1$ \textbf{to} $N$} {
		\For{$j \gets 1$ \textbf{to} $D$} {
			$\zeta^j_i = \text{Softmax}(\hat{o}^j_i/T)$\;
			$o^j_i = \text{one\_hot}(\argmax_k \hat{o}^{jk}_i)$\;
			$o^j_i = \texttt{stop\_gradient}(o^j_i - \zeta^j_i) + \zeta^j_i$\;
		}
		A step of SGD on $\{\hat{o}^j_i\}, \{\mathcal{W}^j\}, \theta$ to reduce $\bigg(v_i - f\bigg((o^1_i)^T \mathcal{W}^1, (o^2_i)^T \mathcal{W}^2, \cdots, (o^D_i)^T \mathcal{W}^D; \theta\bigg)\bigg)^2$
	}
	\caption{An epoch of code learning via Straight-through Estimator with Tempering Softmax.}
	\label{algo:STE_softmax}
\end{algorithm}

%% file: content/exp.tex
\section{Experiments}

In this section we present both real and synthetic experiments to validate our proposed approach. The first set of experiments are based on language modeling task. The language modeling is a fundamental task in NLP, and it can be formulated as predicting the probability over a sequence of words. Models based on recurrent neural networks with word embedding \cite{mikolov2010recurrent,kim2016character} achieve state-of-the-art results, so on which we will base our experiments. The widely used English Penn Treebank \cite{marcus1993building} dataset is used in our experiments, which contains 1M words with vocabulary size of 10K. The training/validation/test split is by convention according to \cite{mikolov2010recurrent}. We utilize standard LSTM \cite{hochreiter1997long} with two different model sizes, which trade-off model size and accuracy. The larger model has word embedding size and LSTM hidden size of 1500, and the number is 200 for the smaller model. By default, $K=50, D=10$ is used in the proposed approach. A temperature schedule, i.e. $T_t = T_0 / (1 + \text{decay\_rate} * t)$, is used to train the code, where $T_0 = 1, \text{decay\_rate}=1$, and $t$ is the iteration number. We first train the model regularly using conventional embedding approach to obtain the embedding vectors, which are used to learn discrete codes. Once the discrete codes are obtained and fixed, we re-train the model with the same architecture and hyper-parameters for the code embedding from scratch.

Table \ref{tab:perfs} shows the performance comparisons between the conventional ``one-hot'' word embeddings against the proposed KD encoding. We presents several variants of the KD encoding schemes, distinguished by the combinations of (1) discrete code learning model and (2) symbol embedding re-learning/re-training model. For the discrete code learning, we have three cases: random assignment, code learned by a linear transformation, and code learned by a LSTM transformation function; the latter two can also be utilized in the symbol embedding re-learning model. Firstly, we observe that the discrete code learning is critical for KD encoding, as random discrete codes produce much worse performance. Secondly, we observe that with appropriate code learning, the test perplexity is similar or better compared to the ``one-hot'' encoding case, while saving 82\%-97\% of embedding parameters.

\begin{table}[t]
	\small
	\centering
	\caption{Comparisons of language modeling in PTB. Test perplexity, embedding size, and compression rate are shown for both small and large model settings. See text for variants of KD encoding.}
	\label{tab:perfs}
	\begin{tabular}{crrrrrr}\Xhline{2\arrayrulewidth}
		& \multicolumn{3}{c}{Small model}                & \multicolumn{3}{c}{Large model}                \\
		& PPL & E. Size & C. Rate & PPL & E. Size & C. Rate \\ \hline
		Conventional          & 114.53    & 2M             & 1                & \textbf{84.04}      & 15M            & 1                \\ \hline
		Random + Linear    & 144.32    & 0.1M  & 0.05             &  103.44      & 0.4M  & 0.033            \\
		Random + LSTM      & 147.13    & 0.37M & 0.185            & 119.62    & 0.63M & 0.042            \\ \hline
		Linear  + Linear & 118.40    & 0.1M  & 0.05             & 87.42     & 0.4M  & 0.033            \\
		Linear  + LSTM   & \textbf{111.13}    & 0.37M & 0.185            & 88.82     & 0.63M & 0.042            \\
		LSTM + Linear    & 117.21    & 0.1M  & 0.05             & \textbf{84.61}     & 0.4M  & 0.033            \\
		LSTM + LSTM      & \textbf{111.31}    & 0.37M & 0.185            & 85.37     & 0.63M & 0.042            \\ \Xhline{2\arrayrulewidth}
	\end{tabular}
\end{table}

\begin{figure}
	\centering
	\begin{subfigure}[b]{0.24\textwidth}
		\includegraphics[width=\textwidth]{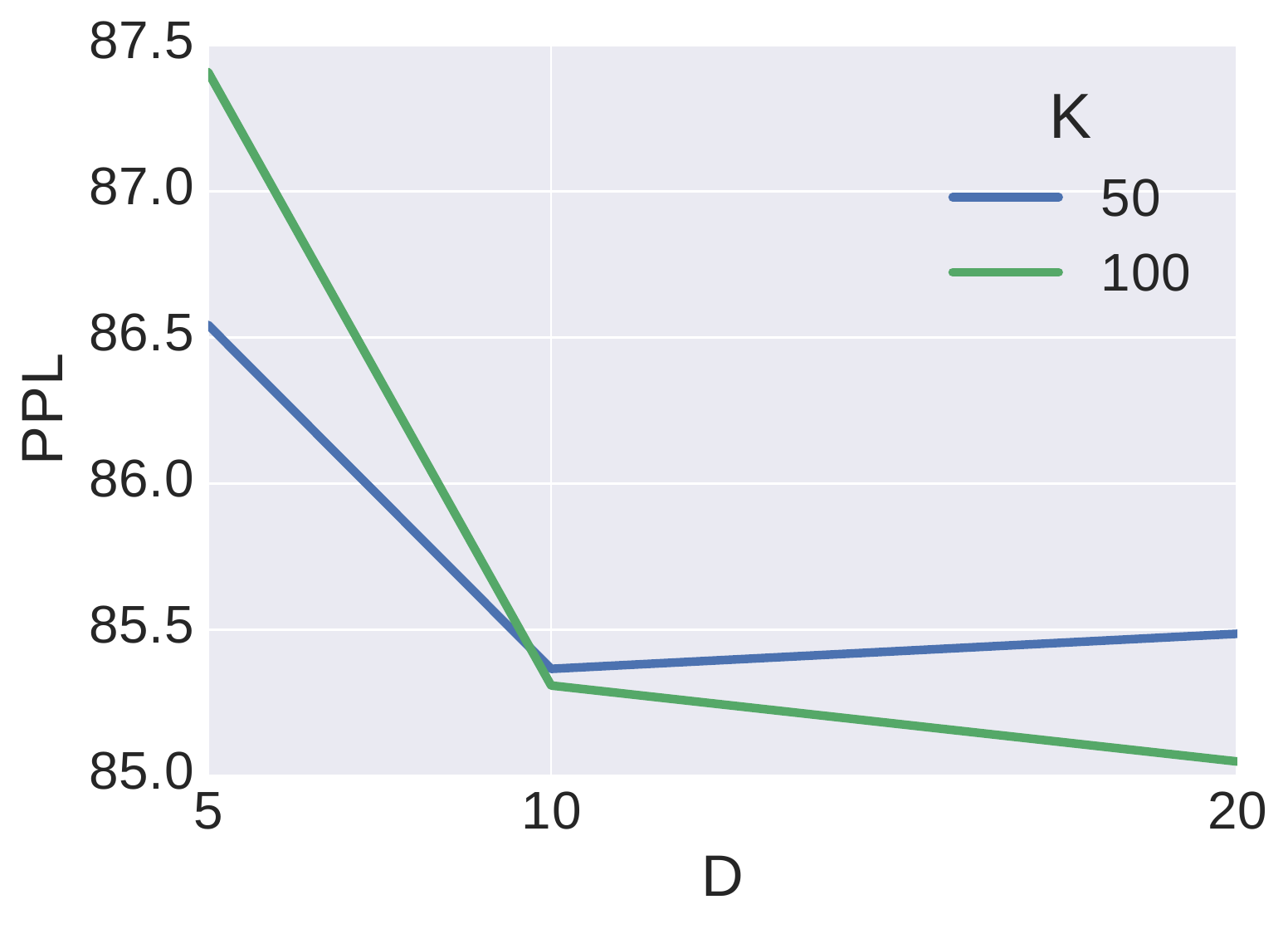}
		\caption{Fix K vary D.}
		\label{fig:fixK_varyD}
	\end{subfigure}
	\begin{subfigure}[b]{0.24\textwidth}
		\includegraphics[width=\textwidth]{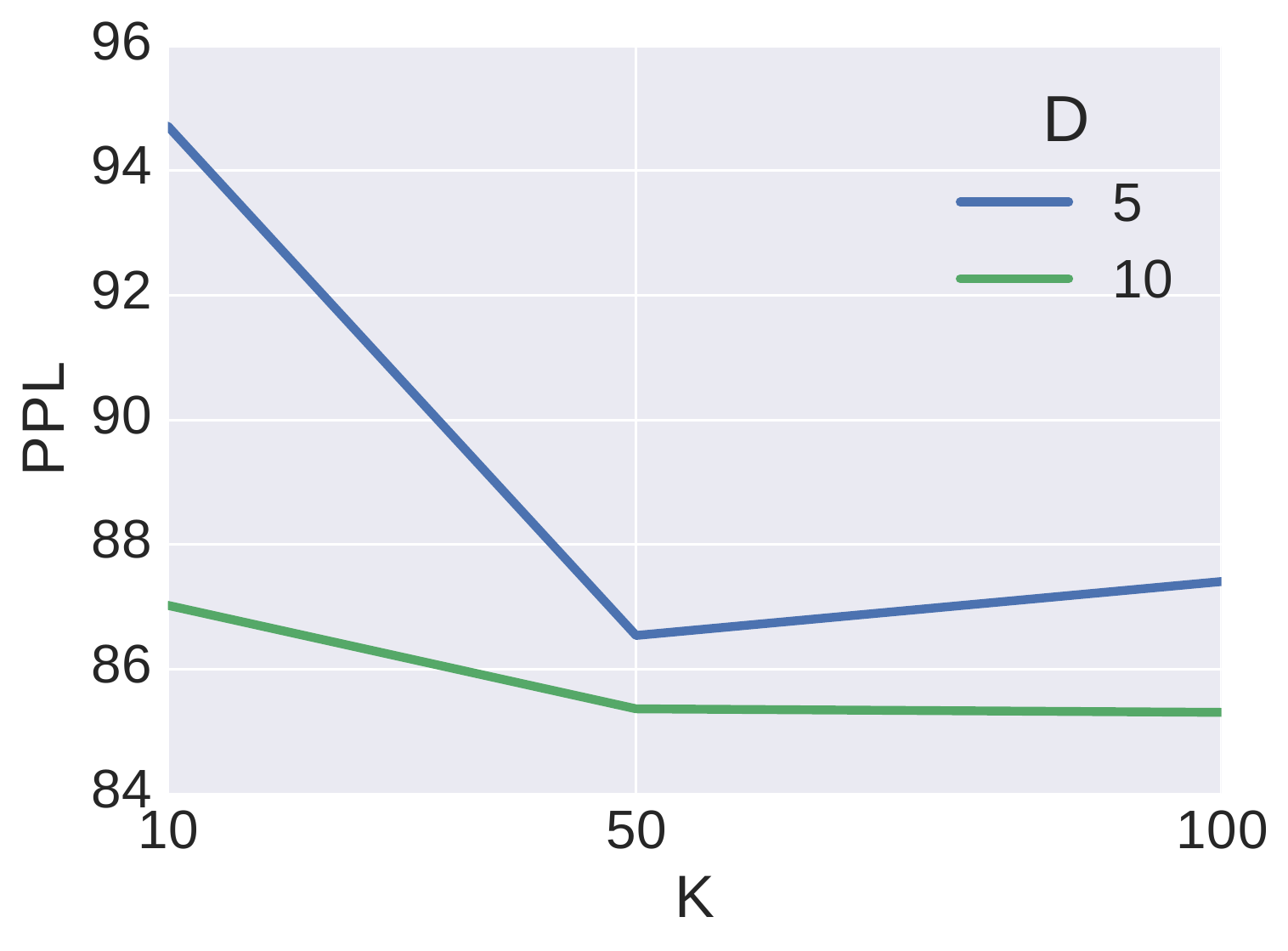}
		\caption{Fix D vary K.}
		\label{fig:fixD_varyK}
	\end{subfigure}
	\begin{subfigure}[b]{0.24\textwidth}
		\includegraphics[width=\textwidth]{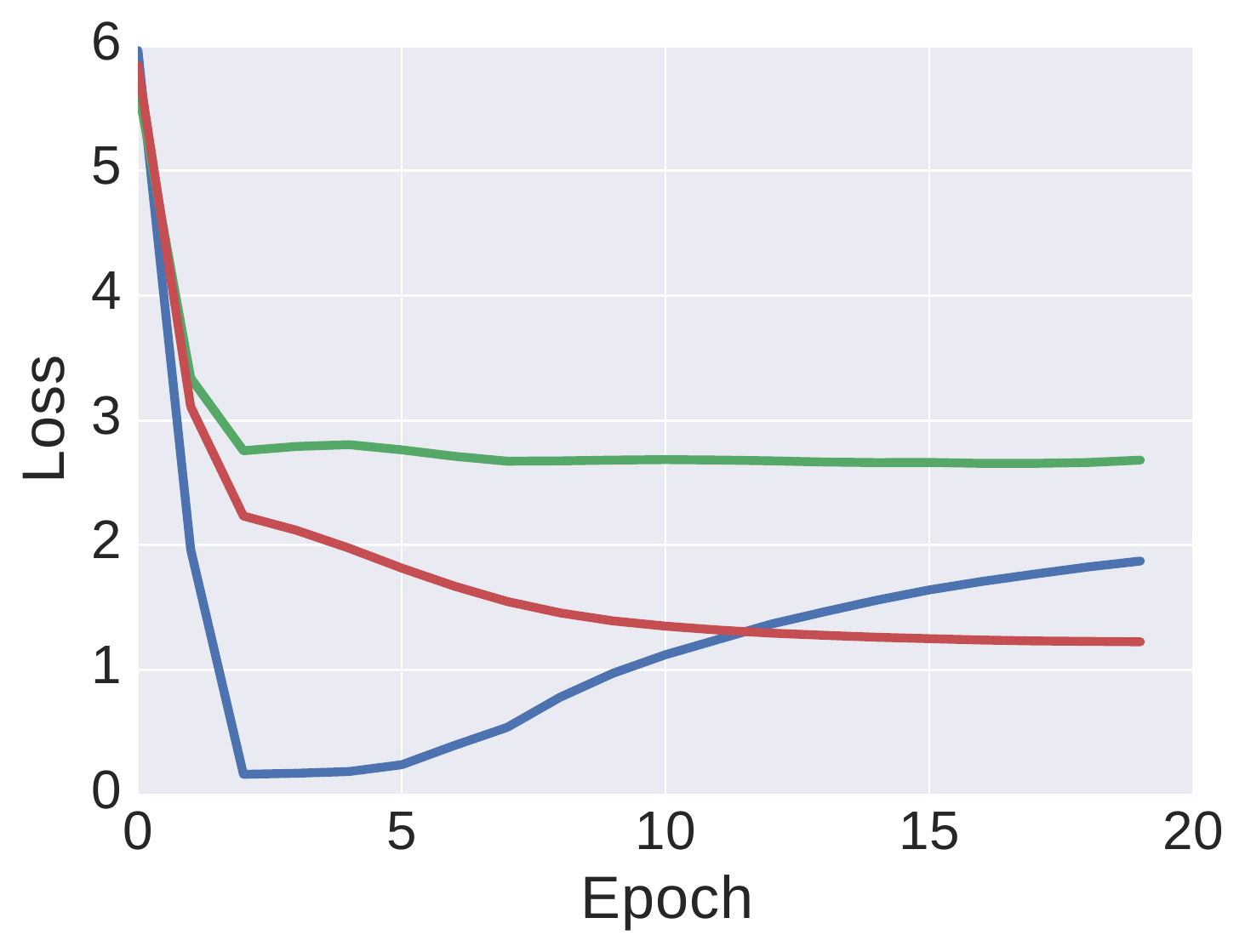}
		\caption{Clustering Loss}
		\label{fig:clustering_loss}
	\end{subfigure}
	\begin{subfigure}[b]{0.24\textwidth}
		\includegraphics[width=\textwidth]{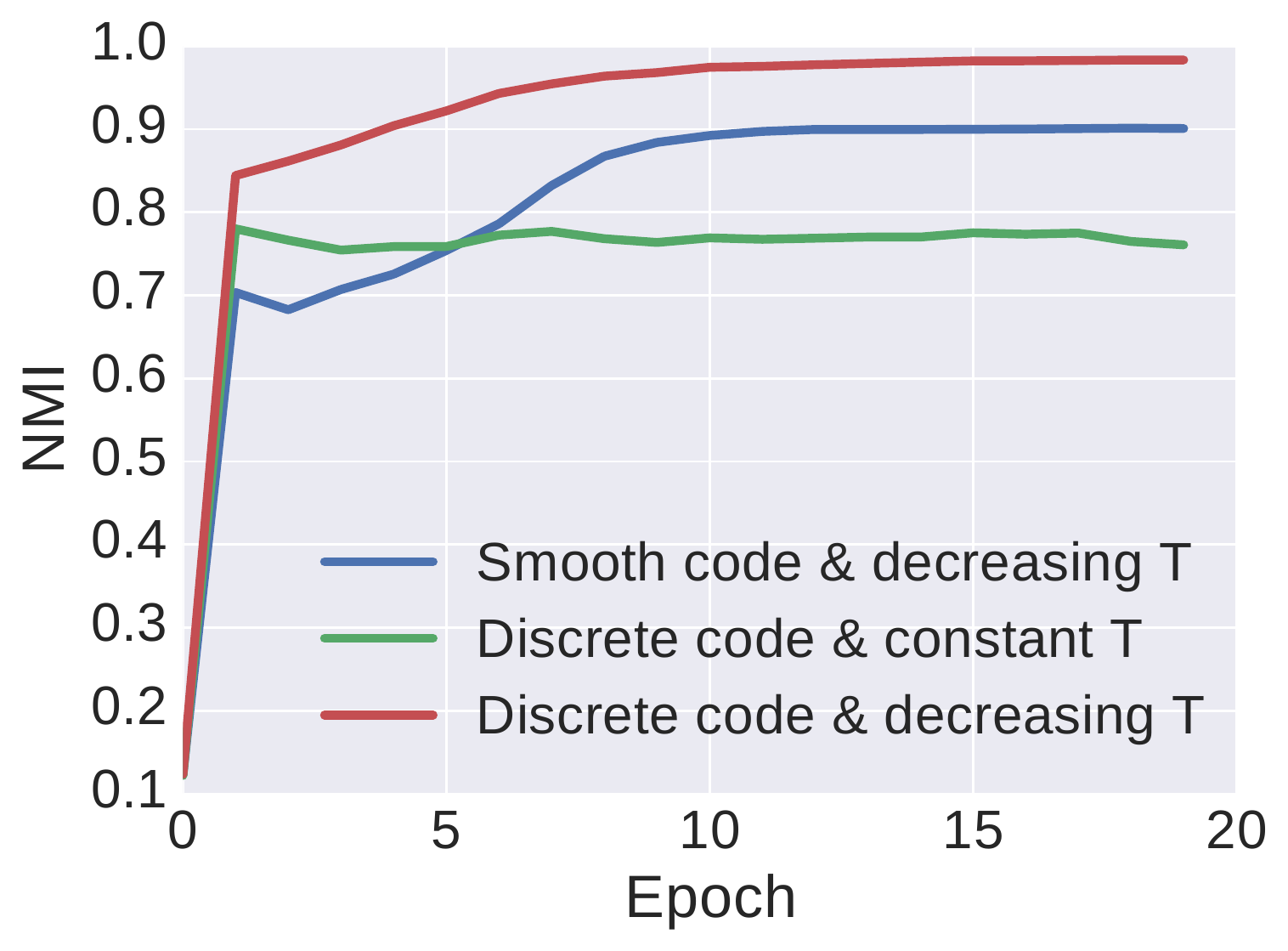}
		\caption{Clustering NMI}
		\label{fig:clustering_nmi}
	\end{subfigure}
	\caption{(a) and (b) are clustering results on synthetic tasks. (c) and (d) are varying K/D on the PTB language modeling task.}\label{fig:clustering_n_kd}

\end{figure}

We also vary the size of $K$ or $D$ and see how they affect the performance. As shown in Figure \ref{fig:fixK_varyD} and \ref{fig:fixD_varyK}, small K or D may harm the performance (even though that $K^D \gg N$ is satisfied), which suggests that the redundant code may be easier to learn.

In order to understand the effects of temperature, and the importance of using discrete code output (i.e., with zero temperature), we create another set of experiments based on the synthetic embedding clusters. We generate 10K nodes that belong to 100 well separated clusters in 10-dimensional space. And $K=100, D=1$ is used, which mimics the K-means clustering problem as each code represents a cluster assignment. Both squared loss and clustering NMI are shown in Figure \ref{fig:clustering_loss} and \ref{fig:clustering_nmi}. We observed that the STE with temperature scheduling is much more effective comparing to its counterparts. When the temperature is kept constant, there are always some percent of codes changing, and the loss as well as NMI converge to a worse local optimal. When a smooth continuous code instead of discrete code is used, we observe that the loss first decreases and then increases. This is due to that only when temperature is small enough, its behavior mimics the discrete code output.

To further inspect the learned code, we use the pre-trained embedding from Glove \cite{pennington2014glove}, which has better coverage and quality than the pre-trained from PTB language modeling. We intentionally use $K=6, D=4$ (code space is 1296) for vocabulary size of 10K, such that the model is forced to collide words. Table \ref{tab:code_demo} show the learned code based on Glove vectors, which demonstrates that similar discrete codes are learned for semantically similar words.

\begin{table}
	\centering
	\caption{Learned code for K=6, D=4 in 10K Glove word embeddings.}
	\label{tab:code_demo}
	\begin{tabular}{ll}\Xhline{2\arrayrulewidth}
		Code    & Words\\ \hline
		3-1-0-3 & up when over into time back off set left open half behind quickly starts\\
		3-1-0-4 & week tuesday wednesday monday thursday friday sunday saturday \\
		3-1-0-5 & by were after before while past ago close soon recently continued meanwhile \\
		3-1-1-1 & year month months record fall annual target cuts      \\\Xhline{2\arrayrulewidth}
	\end{tabular}
\end{table}

%% file: content/related.tex
\section{Related Work}

The idea of using more efficient coding system dates back to information theory, such as error correction code \cite{hamming1950error}, and Hoffman code \cite{huffman1952method}. However, in most embedding techniques such as word embedding \cite{mikolov2013distributed,pennington2014glove}, entity embedding \cite{chen2016entity}, ``one-hot'' encoding is used along with a usually large embedding matrix. Recent work \cite{kim2016character,sennrich2015neural,zhang2015character} explores character or sub-word based embedding model instead of the word embedding model yields some good results. However, in their cases, the chars and sub-words are fixed and given a priori according to the language, thus may have few semantic meanings attached and not available for other data. In contrast, we learn the code assignment function from data, as well as using a fixed length $D$ for the code. 

The compression of neural networks \cite{han2015deep,han2015learning,chen2015compressing} has risen to be an important and hot topic as the size of parameters is too large and becomes a bottleneck for deploying the model to mobile devices. Our work can also be seen as a way to compress the embedding layer in neural networks. Most existing network compression techniques focus on layers that are shared in all data examples, while only one or a few symbols will be utilized in embedding layer at a time in our work.

LightRNN \cite{li2016lightrnn} can be seen as a special case of the proposed KD code, where $K=\sqrt{N}$, $D=2$. Due to the use of a more compact code, code learning is harder and more expensive. We also note that a similar work of encoding embeddings with discrete codes \cite{shu2017compress} is conducted in parallel to ours.


%% file: content/conclusion.tex
\section{Conclusions and Future Work}

In this paper, we propose a novel K-way D-dimensional discrete encoding scheme to replace the ``one-hot" encoding. By adopting the new coding system, the efficiency of parameterization can be significantly improved. Furthermore, the reduction of parameters can also mitigate the over-fitting problem. To learn the semantically meaningful code, we derive a relaxed discrete optimization technique based on SGD. In our experiments of language modeling, the number of free parameters can be reduced by 97\% while achieving similar or better performance. We are currently working on improving the on-the-fly KD code learning along with the given tasks, where the symbol embeddings are not given beforehand. 

%% file: main.bbl
\begin{thebibliography}{20}
\providecommand{\natexlab}[1]{#1}
\providecommand{\url}[1]{\texttt{#1}}
\expandafter\ifx\csname urlstyle\endcsname\relax
  \providecommand{\doi}[1]{doi: #1}\else
  \providecommand{\doi}{doi: \begingroup \urlstyle{rm}\Url}\fi

\bibitem[Bengio et~al.(2013)Bengio, L{\'e}onard, and
  Courville]{bengio2013estimating}
Yoshua Bengio, Nicholas L{\'e}onard, and Aaron Courville.
\newblock Estimating or propagating gradients through stochastic neurons for
  conditional computation.
\newblock \emph{arXiv preprint arXiv:1308.3432}, 2013.

\bibitem[Bordes et~al.(2013)Bordes, Usunier, Garcia-Duran, Weston, and
  Yakhnenko]{bordes2013translating}
Antoine Bordes, Nicolas Usunier, Alberto Garcia-Duran, Jason Weston, and Oksana
  Yakhnenko.
\newblock Translating embeddings for modeling multi-relational data.
\newblock In \emph{Advances in neural information processing systems}, pages
  2787--2795, 2013.

\bibitem[Chen et~al.(2016)Chen, Tang, Sun, Chen, and Zhang]{chen2016entity}
Ting Chen, Lu-An Tang, Yizhou Sun, Zhengzhang Chen, and Kai Zhang.
\newblock Entity embedding-based anomaly detection for heterogeneous
  categorical events.
\newblock In \emph{Proceedings of the Twenty-Fifth International Joint
  Conference on Artificial Intelligence}, pages 1396--1403. AAAI Press, 2016.

\bibitem[Chen et~al.(2015)Chen, Wilson, Tyree, Weinberger, and
  Chen]{chen2015compressing}
Wenlin Chen, James Wilson, Stephen Tyree, Kilian Weinberger, and Yixin Chen.
\newblock Compressing neural networks with the hashing trick.
\newblock In \emph{International Conference on Machine Learning}, pages
  2285--2294, 2015.

\bibitem[Hamming(1950)]{hamming1950error}
Richard~W Hamming.
\newblock Error detecting and error correcting codes.
\newblock \emph{Bell Labs Technical Journal}, 29\penalty0 (2):\penalty0
  147--160, 1950.

\bibitem[Han et~al.(2015{\natexlab{a}})Han, Mao, and Dally]{han2015deep}
Song Han, Huizi Mao, and William~J Dally.
\newblock Deep compression: Compressing deep neural networks with pruning,
  trained quantization and huffman coding.
\newblock \emph{arXiv preprint arXiv:1510.00149}, 2015{\natexlab{a}}.

\bibitem[Han et~al.(2015{\natexlab{b}})Han, Pool, Tran, and
  Dally]{han2015learning}
Song Han, Jeff Pool, John Tran, and William Dally.
\newblock Learning both weights and connections for efficient neural network.
\newblock In \emph{Advances in Neural Information Processing Systems}, pages
  1135--1143, 2015{\natexlab{b}}.

\bibitem[Hochreiter and Schmidhuber(1997)]{hochreiter1997long}
Sepp Hochreiter and J{\"u}rgen Schmidhuber.
\newblock Long short-term memory.
\newblock \emph{Neural computation}, 9\penalty0 (8):\penalty0 1735--1780, 1997.

\bibitem[Huffman(1952)]{huffman1952method}
David~A Huffman.
\newblock A method for the construction of minimum-redundancy codes.
\newblock \emph{Proceedings of the IRE}, 40\penalty0 (9):\penalty0 1098--1101,
  1952.

\bibitem[Jang et~al.(2016)Jang, Gu, and Poole]{jang2016categorical}
Eric Jang, Shixiang Gu, and Ben Poole.
\newblock Categorical reparameterization with gumbel-softmax.
\newblock \emph{arXiv preprint arXiv:1611.01144}, 2016.

\bibitem[Kim et~al.(2016)Kim, Jernite, Sontag, and Rush]{kim2016character}
Yoon Kim, Yacine Jernite, David Sontag, and Alexander~M Rush.
\newblock Character-aware neural language models.
\newblock In \emph{Proceedings of the Thirtieth AAAI Conference on Artificial
  Intelligence}, pages 2741--2749. AAAI Press, 2016.

\bibitem[Li et~al.(2016)Li, Qin, Yang, Hu, and Liu]{li2016lightrnn}
Xiang Li, Tao Qin, Jian Yang, Xiaolin Hu, and Tieyan Liu.
\newblock Lightrnn: Memory and computation-efficient recurrent neural networks.
\newblock In \emph{Advances in Neural Information Processing Systems}, pages
  4385--4393, 2016.

\bibitem[Maddison et~al.(2016)Maddison, Mnih, and Teh]{maddison2016concrete}
Chris~J Maddison, Andriy Mnih, and Yee~Whye Teh.
\newblock The concrete distribution: A continuous relaxation of discrete random
  variables.
\newblock \emph{arXiv preprint arXiv:1611.00712}, 2016.

\bibitem[Marcus et~al.(1993)Marcus, Marcinkiewicz, and
  Santorini]{marcus1993building}
Mitchell~P Marcus, Mary~Ann Marcinkiewicz, and Beatrice Santorini.
\newblock Building a large annotated corpus of english: The penn treebank.
\newblock \emph{Computational linguistics}, 19\penalty0 (2):\penalty0 313--330,
  1993.

\bibitem[Mikolov et~al.(2010)Mikolov, Karafi{\'a}t, Burget, {\v{C}}ernock{\`y},
  and Khudanpur]{mikolov2010recurrent}
Tom{\'a}{\v{s}} Mikolov, Martin Karafi{\'a}t, Luk{\'a}{\v{s}} Burget, Jan
  {\v{C}}ernock{\`y}, and Sanjeev Khudanpur.
\newblock Recurrent neural network based language model.
\newblock In \emph{Eleventh Annual Conference of the International Speech
  Communication Association}, 2010.

\bibitem[Mikolov et~al.(2013)Mikolov, Sutskever, Chen, Corrado, and
  Dean]{mikolov2013distributed}
Tomas Mikolov, Ilya Sutskever, Kai Chen, Greg~S Corrado, and Jeff Dean.
\newblock Distributed representations of words and phrases and their
  compositionality.
\newblock In \emph{Advances in neural information processing systems}, pages
  3111--3119, 2013.

\bibitem[Pennington et~al.(2014)Pennington, Socher, and
  Manning]{pennington2014glove}
Jeffrey Pennington, Richard Socher, and Christopher Manning.
\newblock Glove: Global vectors for word representation.
\newblock In \emph{Proceedings of the 2014 conference on empirical methods in
  natural language processing}, pages 1532--1543, 2014.

\bibitem[Sennrich et~al.(2015)Sennrich, Haddow, and Birch]{sennrich2015neural}
Rico Sennrich, Barry Haddow, and Alexandra Birch.
\newblock Neural machine translation of rare words with subword units.
\newblock \emph{arXiv preprint arXiv:1508.07909}, 2015.

\bibitem[Shu(2017)]{shu2017compress}
Hideki Shu, Raphael;~Nakayama.
\newblock Compressing word embeddings via deep compositional code learning.
\newblock \emph{https://arxiv.org/abs/1711.01068}, 2017.

\bibitem[Zhang et~al.(2015)Zhang, Zhao, and LeCun]{zhang2015character}
Xiang Zhang, Junbo Zhao, and Yann LeCun.
\newblock Character-level convolutional networks for text classification.
\newblock In \emph{Advances in neural information processing systems}, pages
  649--657, 2015.

\end{thebibliography}
